# Occam's Ghost

by Peter Kövesarki

**Abstract**

This article applies the principle of Occam's Razor to non-parametric model building of statistical data, by finding a model with the minimal number of bits, leading to an exceptionally effective regularization method for probability density estimators. The idea comes from the fact that likelihood maximization also minimizes the number of bits required to encode a dataset. However, traditional methods overlook that the optimization of model parameters may also inadvertently play the part in encoding data points. The article shows how to extend the bit counting to the model parameters as well, providing the first true measure of complexity for parametric models. Minimizing the total bit requirement of a model of a dataset favors smaller derivatives, smoother probability density function estimates and most importantly, a phase space with fewer relevant parameters. In fact, it is able prune parameters and detect features with small probability at the same time. It is also shown, how it can be applied to any smooth, non-parametric probability density estimator.

## 1 Introduction

Occam's razor is a principle, perhaps more than two millennia old, stated by a 14th century philosopher monk William Ockham. Often stated as a rule to accept those conjectures as true that are the simplest and explain a set of phenomena thoroughly. Still, a better wording came already from Ptolemy[1], a thousand year earlier as

> We consider it a good principle to explain the phenomena by the simplest hypothesis possible.

This earlier statement is better in the sense, that it reflects the long struggle of scientists, that truth is exceptionally difficult to be found. With time, new information may reveal itself, that contradicts the previous best hypothesis and a new hypothesis must be formulated. Although it is good to see that Occam's Razor is not prone to itself, a similar hypothesis has never been properly formulated to mathematical model building. As anyone who has fitted regression curves or maximum likelihood models to their data could have seen, that a more complex model can result a better fit or a higher likelihood. So where is the tradeoff between model complexity and better data description?

Mathematical fitting might be an ill-posed problem, having multiple solutions satisfying the same fitting criteria, so one ought to implement a form of regularization or include Bayesian priors not only to have a smoother or more preferred solution, but also to avoid what is called overfitting. This latter is a typical plague of statistical methods, where the fit/regression either tends to learn the individual data points or the regression turns into interpolation instead of having predictive power for unseen data. A way to handle overtraining issues is to fit a part of the available data, and check wether the same fitting criteria gives comparably optimal results for the unseen part, that was not used during the optimization[7]. Without explicit testing data, one has to revert to estimate the fit uncertainty by using the existing data[11][7] with bootstrap[5], cross validation or generating synthetic data. Various recipes exist to mitigate the uncertainty, like favoring models with fewer degree of freedom, pruning the existing model or favoring smoother curves. However, there are no singled out measure to optimize, equation or differential equation to solve, algorithm to follow that produces the best result for everyone. There are rules of thumb, best practices, collection of criteria, learnt bit by bit from the tests. Nevertheless, in most situations one just want something that works, without having to worry about smoothness or overtraining.

---

1. https://en.wikipedia.org/wiki/Occam%27s_razor#Formulations_before_William_of_Ockham

One can agree, that for data coming from an unknown distribution it is not possible to reconstruct the true distribution. With more data, more features reveal themselves, finer details can be seen. Hence one both expects the fit function to describe the unknown data well, and also be able to reveal new features when the new data is added. In order to achieve this in general, one needs to retain all the previously used data that was used for the fit, add the new data and start the modeling again. Doesn't this contradict with Occam's razor? Retaining all the data instead of a simple model seems to be bad deal. Still, one can argue that it is enough to retain a simple model, from which one can accurately reconstruct the data. This shows how coding theory and probabilistic modeling are connected, and it is not a new concept. Sparse learning and auto-encoders are related, not uncommon methods, but many have used compression ratios as fitness criterium or utilized the Kullback-Liebler divergence[12], which measures expected number of extra bits required per binary digit for a data point, if other than true distribution is used to transcode it. These latter two methods already reveal that there is a fundamental disagreement between Occam's razor and parametric models, such as that data and model parameters are typically built of real numbers, with infinite number of bits. Therefore each model encompassing a real parameter is already supposed to be infinitely complex. How can we compare models then, and select the simplest?

## 2　Tracking of Precision

One answer to the problem of infinite bits in models with real numbers is not using infinitely precise numbers, at least not directly. We can incorporate precision into model building, and use it as a measure of both model complexity and model fitness. The goal is to reconstruct the dataset $\{x_{\mu i}\}$ with a given precision $\Delta x_\mu > 0$, which later can be taken in the $\Delta x_\mu \to 0$ limit. To encode this data to $u_\nu$ we need a modeling function, a map $x_\nu = G_\nu(u_\mu, m_k)$ with parameters $m_k$. In order for it to work properly, it needs to be bijective, and for effective encoding it is better to be constrained such as $u_\nu \in [0...1]$, so binary coding of $u_\nu$ is most efficient. These requirements make seem $G$ as a one dimensional inverse cumulative distribution function in the $\nu$ direction, though that is not a goal, $u$ might be in any other range than $[0...1]$. An additional requirement, for the sake of simplicity is differentiability, so a small change in $x$ results in a small change in $u$, and as we will see, the same need to be required for the continuous parameters $m_k$. For convenience reasons, we will use the $u \to x$ inverse of $G$, with the same parameterization, so $u_\nu = F_\nu(x_\mu, m_k)$ as it is more similar to cumulative distribution functions.

This similarity to cumulative distribution functions arises from the Jacobi determinant. A coordinate transformation on random variables alters the probability density function via the Jacobi determinant[9], as $\text{Prob}(x) = \text{Prob}(u(x)) \left| \det\left(\frac{\partial u}{\partial x}\right) \right|$. However, for a transformation where $u$ becomes uniformly distributed, $\text{Prob}(u)$ is a constant and $\left| \det\left(\frac{\partial u}{\partial x}\right) \right|$ becomes proportional to $\text{Prob}(x)$. The exact proportion depends on the boundaries of $u$, the volume of where $u$ is understood.

We are interested in $\Delta u_{\nu i}$ and $\Delta m_k$ to calculate number of bits required to reconstruct the dataset, with the given precision, with the model $F$, with parameters $m_k$. That total number of bits is proportional to

$$Q = -\sum_{\nu i} \log \Delta u_{\nu i} - \sum_k \log \Delta m_k \tag{1}$$

For $u_{\nu i}$ it is straightforward to use the truncating precision $\Delta u_{\nu i}$, as the $u_{\nu i}$ variable is confined to a range, but $-\log \Delta m_k$ counts only the number of bits after the decimal point of $m_k$. However, we are interested in the behavior of the model in the limit where $\Delta m_k \to 0$. Finite number of bits, related to model building decisions are also omitted in this approximation.

In order to maintain a $\Delta x_{\mu i}$ precision, we must track all the possible sources that contribute to $x_{\mu i}$, and these include not only that with new data $m_k$ changes, but also that $u_{\nu i}$ is object to change. From the total derivative of $F$

$$\text{d}u_{\nu i} = \frac{\partial F_\nu(x_{\lambda i}, m_k)}{\partial x_\mu} \text{d}x_{\mu i} + \frac{\partial F_\nu(x_{\lambda i}, m_l)}{\partial m_k} \text{d}m_k \tag{2}$$

we can express the change in $x_{\mu i}$ by the changes in $u_{\nu i}$ and $m_k$:

$$\mathrm{d}x_{\mu i} = \left(\frac{\partial F_{\nu i}}{\partial x_\mu}\right)^{-1} \mathrm{d}u_{\nu i} - \left(\frac{\partial F_{\nu i}}{\partial x_\mu}\right)^{-1} \left(\frac{\partial F_\nu(x_{\lambda i}, m_k)}{\partial m_l}\right) \mathrm{d}m_l \quad (3)$$

Now the task is to determine the smallest number of bits, or largest $\Delta u_{\nu i}$ and $\Delta m_k$ ranges, within which $u_{\nu i}$ and $m_k$ can be truncated, but still causing changes smaller than $\Delta x_{\mu i}$. In most cases we are going to treat $\Delta x_{\mu i}, \Delta u_{\nu i}$ and $\Delta m_k$ as non-zero, positive parameters. In some cases though, their sign matters, and then it is indicated in one form or another.

## 2.1 The Baseline Case

We want to find at which $\Delta u_{\nu i}$ and $\Delta m_k$ precision can we modify $u_{\nu i}$ and $m_k$ to keep the reconstructed $x_{\mu i}$ within $\Delta x_{\mu i}$ precision. We must assume, that within this precision the parameters could be anything and they are independent from each other. This is because once the variables are truncated, we can't define them more precisely anymore, and post-select a set of $u_{\nu i}$ and $m_k$ that reconstructs $x_{\mu i}$ with the desired precision. In other words, the assumption is, that after truncation, the next bits of the numbers are not predictable. We should also not encode the $x_{\mu i}$ into the auxiliary $\Delta u_{\nu i}$ and $\Delta m_k$ variables, those are strictly there to track the number of bits of the model. A lesser problem is, that after the truncation, there are further bits of $u_{\nu i}$ and $m_k$, which when included, should not modify $x_{\mu i}$ outside $\Delta x$. Also, the future modifications of the model parameters are unknown, the truncation means replacing $x$ with something within the range $[x - \Delta x, x + \Delta x]$, without being able to reconstruct the original $x$. Therefore the problem is not a general linear programming task, that finds a set of perturbations that still allow small $\Delta x$, but rather finding the worst case scenario at which $x_{\mu i}$ can still be reconstructed with the given precision. With this in mind we can write that our goal is to find $\Delta m_k > 0$ and $\Delta u_{\nu i} > 0$ where

$$\Delta x = \Delta x_{\mu i} \geqslant \left|\left(\frac{\partial F_{\nu i}}{\partial x_\mu}\right)^{-1}\right| \Delta u_{\nu i} + \left|\left(\frac{\partial F_{\nu i}}{\partial x_\mu}\right)^{-1} \left(\frac{\partial F_\nu(x_{\lambda i}, m_k)}{\partial m_l}\right)\right| \Delta m_l > 0 \quad (4)$$

Where the absolute values are taken element-wise for the resulting matrices. Generally, we want a model $F$ which reconstructs the data with the required precision, but uses the minimal number of bits. We can see, that typical models have as many $\Delta u_{\nu i}$ degrees of freedom as there are $\Delta x_{\mu i}$-s. Therefore, for any fixed set of $\Delta m_k$ and $\Delta x_{\mu i}$ for which the inequality is satisfied, we are always allowed to increase the corresponding set of $\Delta u_{\nu i}$ (the same event index $i$), and decrease eq. 1 up until the point that the equality is fulfilled. This raises the possibility to explicitly express $\Delta u_{\nu i}$ as a function of $\Delta x_{\mu i}$ and $\Delta m_k$:

$$\Delta u_{\nu i} = \left(\left|\left(\frac{\partial F_{\nu i}}{\partial x_\mu}\right)^{-1}\right|\right)^{-1} \Delta x_{\mu i} - \left(\left|\left(\frac{\partial F_{\nu i}}{\partial x_\mu}\right)^{-1}\right|\right)^{-1} \cdot \left|\left(\frac{\partial F_{\alpha i}}{\partial x_\mu}\right)^{-1} \left(\frac{\partial F_\alpha(x_{\lambda i}, m_l)}{\partial m_k}\right)\right| \Delta m_k \quad (5)$$

Without the element-wise absolute values. eq. 5 would look similar to a total derivative, except for the minus sign in front of the $\Delta m_k$ terms. The equation takes a bit simpler form when derived from the $x = G(u, m)$ inverse function, the minus sign remains nevertheless, since the inequality 4 on $\Delta x$ precision still has to be applied. The form applying $F(x, m)$ instead of $G(u, m)$ is preferable, as in most cases one wants to fit a sum of weighted PDFs, which is rather complicated using the inverse cumulative distributions.

Eq. 5 can be readily substituted into the objective function in eq. 1, making it a function of $m_k$ and $\Delta m_k$ only (as $\Delta x_{\mu i}$ is treated as a constant), greatly reducing the number of parameters and making it much more feasible for optimizations.

In case $\Delta x_{\mu i}$ are all the same $\Delta x$, we can isolate $\Delta x$ within $Q$ by substituting $\Delta x$ with 1, $\Delta m_k$ with $\frac{\Delta m_k}{\Delta x}$ and subtracting $(N_e N_{\mathrm{dim}} - N_p)\log(\Delta x)$ from $Q$, where $N_e$ is the number of data points, $N_{\mathrm{dim}}$ the number of dimensions and $N_p$ are the number of parameters. Thus removing the $(N_e N_{\mathrm{dim}} - N_p)\log(\Delta x)$ part, $Q$ may depend on the $m_k$ parameters and the $\frac{\Delta m_k}{\Delta x}$ ratios. However, this also means that in the $\Delta x \to 0$ limit, only models with the same number of parameters should be compared.

## 2.2 General Case

As we are used to in probability theory, we can try to substitute the $\frac{\partial F_\nu}{\partial x_\mu}$ derivatives with probability density functions. As $Q$ contains $\sum \log(\Delta u_{\nu i})$, we can say that it is enough to calculate the required $\Delta u^n$ volume instead of the individual $\Delta u_{\nu i}$ precisions, and trade the precision of one dimension for others', but keep $\prod \Delta u_{\nu i} = \Delta u_i^n$ the constant. An other argument for this case is, that when we want to truncate a number, we look for the closest dot on a fixed digit grid. Hence for a given precision, we need to ensure that the volume is large enough to contain at least one dot of the grid. Without the $\frac{\partial F}{\partial m_k} \mathrm{d} m_k$ terms in the total derivative in eq. 2, we are left with the less problematic $\frac{\partial F}{\partial x_\mu} \mathrm{d} x_\mu$ terms. The term $\frac{\partial F}{\partial x}$ turns the cuboid $\Delta x_{\mu i}$ into an $n$-parallelotope of $\Delta u_{\nu i}$, and so the inverse turns the $\Delta u_{\nu i}$ cuboid into an $n$-parallelotope of $\Delta x_{\mu i}$. The volume encompassed by $\Delta u_{\nu i}$ can be expressed with the determinant of the transformation as $\left|\det \frac{\partial F_\nu}{\partial x_\mu}\right| \prod \Delta x_{\mu i}$.

As the goal is keeping a given precision on the reconstructed data set $x_{\mu i}$, we should check how the $\Delta x_{\mu i}$ cuboid transforms with the help of eq. 2. The image of the $\Delta x_{\mu i}$ in $u$-space is an $n$-parallelotope, whose vertices in the simplest case, without the $\mathrm{d}m$ perturbations are at $\frac{\partial F_\nu}{\partial x_\mu} \Delta x_{\mu i}$, with $\Delta x_{\mu i}$ taking up all the values of $\pm \Delta x$ for a given $i$. A given nonzero $\Delta m_a$ perturbation shifts the points of the $n$-parallelotope in $u$-space with the vector $\frac{\partial F}{\partial m_a} \Delta m_a$. Those points that are shifted out from the unperturbed boundaries of the parallelotope, due to the independent[2] nature of $\Delta u_{\nu i}$ and $\Delta m_k$, would be outside the cuboid of the allowed precision. The volume of this veto-region is shrinking the allowed volume of $u$ that can be used in encoding the data. The volume is like a shadow of the unperturbed $n$-parallelotope, the area of its orthogonal projection in the direction of $\frac{\partial F_\nu}{\partial m_a}$, multiplied with the length $\left|\frac{\partial F_\nu}{\partial m_a} \Delta m_a\right|$. The same is be valid for all the parameters $m_k$, and also for perturbations in the $-\Delta m_k$ directions. Thus, the available volume around $u_{\nu i}$ has a maximum value, the familiar $\left|\det \frac{\partial F_\nu}{\partial x_\mu}\right| \prod \Delta x_{\mu i}$, from which the veto-volumes have to be subtracted. To calculate those, the procedure is replacing a column of the $\frac{\partial F_\nu}{\partial x_\mu}$ matrix with $\frac{\partial F_\nu}{\partial m_k}$, now representing a face of the $n$-parallelotope and creating a new one with a new edge $\frac{\partial F_\nu}{\partial m_k}$. The absolute value of the determinant of this resulting matrix, multiplied by the corresponding $\Delta x_{\mu i}$ and $\Delta m_k$ (and the occasional $2^n$ factor representing $\pm \Delta x_{\mu i}$ instead of the $[0, \Delta x_{\mu i}]$ region) will give the veto-volume for one face of the parallelotope with one $\frac{\partial F_\nu}{\partial m_k}$ direction, so this has to be done for all the columns of the $\frac{\partial F_\nu}{\partial x_\mu}$ matrix, replaced with all the $\frac{\partial F_\nu}{\partial m_k}$ shifts. This exercise was done in order to show that the maximal volume one can use around $u_{\nu i}$ is determined by the Jacobi determinant, proportional to $\left|\det \frac{\partial F_\nu}{\partial x_\mu}\right|$, and the perturbations can only decrease the available volume.

We can simplify this calculation further, though. As seen in the previous paragraph, the maximal value of the $\prod \Delta u$ volume is $\left|\det \frac{\partial F_\nu}{\partial x_\mu}\right| \prod \Delta x$, while the further subtracted terms only differ by one row from the $\frac{\partial F_\nu}{\partial x_\mu}$ matrix. For a single $\frac{\partial F_\nu}{\partial m_a} \Delta m_a$ term this is equivalent of shifting a single $\frac{\partial F_\nu}{\partial x_b} \Delta x_{bi}$ edge of the maximal $\prod \Delta u$ $n$-parallelotope in a direction that decreases the volume. In other words, the new edge is $\frac{\partial F_\nu(x_{\lambda i}, m_k)}{\partial x_b} \Delta x_{bi} + \frac{\partial F_\nu(x_{\lambda i}, m_l)}{\partial m_a} \Delta m_a$, where $\Delta m_a$ is restricted to values that decrease the absolute value of the determinant of the new matrix. To achieve this decrease, we must not simply find the direction that decreases the length of $\frac{\partial F_\nu(x_{\lambda i}, m_k)}{\partial x_b} \Delta x_{bi}$, but we need a direction that is perpendicular to all columns of $\frac{\partial F_\nu}{\partial x_\mu}$ where $\mu \neq b$. To obtain this, an obvious choice is the $(n-1)$-form exterior product created from the $\mu \neq b$ vectors, but that is computationally very costly, and also computes the unneeded length of this vector. A simpler approach is orthogonalization of the $\frac{\partial F_\nu}{\partial x_b}$ with respect to the $\frac{\partial F_\nu}{\partial x_\mu}$ $\mu \neq b$ vectors, which we can denote as $\frac{\partial F_\nu}{\partial x_{b\perp}}$. Therefore the sign of $\Delta m_a$ should be the one that gives $\text{sign}(\Delta m_a) = -\text{sign}\left(\frac{\partial F_\nu(x_{\lambda i}, m_k)}{\partial x_{b\perp}} \frac{\partial F_\nu(x_{\lambda i}, m_l)}{\partial m_a} \Delta x_{bi}\right)$, an angle larger than $\frac{\pi}{2}$. The maximal value of $|\Delta m_a|$, is a bit more problematic. It can't be too large, since a large enough $\Delta m_a$ could increase the absolute volume, independently of its sign. The goal is to fulfill the constraint, that a $u_{\nu i}$ point is not shifted out of the allowed $\prod \Delta x$ volume, therefore only those $\Delta m_a$ values are allowed that do not flip or annul the the sign of the determinant, and guarantees that $\text{sign}\left(\det\left(\frac{\partial F_\nu(x_{\lambda i}, m_l)}{\partial x_\mu}\right)\prod \Delta x_{\mu i}\right) = \text{sign}\left(\det\left(\frac{\partial F_\nu(x_{\lambda i}, m_l)}{\partial x_\mu}\Delta x_{\mu i} + \frac{\partial F_\nu(x_{\lambda i}, m_l)}{\partial m_k}\Delta m_k\right)\right)$. The easiest way to identify this is to check wether the $\frac{\partial F_\nu(x_{\lambda i}, m_l)}{\partial x_\mu}\Delta x_{\mu i} + \frac{\partial F_\nu(x_{\lambda i}, m_l)}{\partial m_k}\Delta m_k$ passes the plane perpendicular to $\frac{\partial F_\nu(x_{\lambda i}, m_k)}{\partial x_{b\perp}}$.

---

2. Independency in this case are the assumptions, that first, the truncation of the $u_{\nu i}$ and $m_k$ are unpredictable, and second, that the truncation of one does not assume the truncation of the other. The precision of $x_{\mu i}$ could be maintained by many sets of $u_{\nu i}$ and $m_k$, but we are looking for a set where large deviations by truncations are allowed, and where we can also interchange the order of truncation. With finite $\Delta m_k$, when the shift to $m_k$ is chosen independently to the shift to $u_{\nu i}$, there exists $m_k$ deviations that shifts the reconstructed $x_{\mu i}$ from the allowed precision region, hence those $u_{\nu i}$ deviations are disallowed. Theoretically, methods where the precision of $m_k$ could be sacrificed to the precision of $u_{\nu i}$ is possible (similar to keeping the $\prod \Delta u_{\nu i}$ volume constant), but due to the largely different number of $m_k$ and $u_{\nu i}$ parameters, their groupings are rather complicated. In the process of fitting, a single $m_k$ parameter is affected by many $x_{\mu i}$ data points, so the $m_k$ truncation shifts many reconstructed $x_{\mu i}$, as opposed to the deviation by truncation of a single $u_{\nu i}$ affects only a single reconstructed $x_{\mu i}$.

So in summary, a simple way to approximate the available volume around $u_{\nu i}$ is determining the sign $s(i, \mu, k) \in \{-1, 1\}$ for each pair of $\mu$ data vector index and $k$ parameter index, for every $i$ data index as

$$s(i, \mu, k) = \text{sign}\left(\frac{\partial F_\nu(x_{\lambda i}, m_l)}{\partial x_{b\perp}}\frac{\partial F_\nu(x_{\lambda i}, m_l)}{\partial m_a}\right)$$

Where $\frac{\partial F_\nu(x_{\lambda i}, m_k)}{\partial x_{b\perp}}$ is a vector created from $\frac{\partial F_\nu(x_{\lambda i}, m_k)}{\partial x_b}$ by orthogonalizing it to every $\frac{\partial F_\nu(x_{\lambda i}, m_k)}{\partial x_\mu}$ where $\mu \neq b$. With the help of $s(i, \mu, k)$ and eq. ? we can create a matrix, whose determinant gives the volume of the allowed $\prod \Delta u$ volume in the lowest order approximation

$$V_{\Delta u_i} = \prod \Delta u_{\nu i} = \left|\det\left(\frac{\partial F_\nu(x_\lambda, m_l)}{\partial x_\mu}|\Delta x_{\mu i}| - \sum_k s(i, \mu, k)\frac{\partial F_\nu(x_\lambda, m_l)}{\partial m_k}|\Delta m_k|\right)\right| \qquad (6)$$

assuming that $\Delta m_k$ are small enough and don't change the sign of the unperturbed determinant. In eq. 6 the matrix has $\mu$ and $\nu$ as indices, no Einstein summation is done on $\mu$ as previously, only the index $k$ is summed up. The absolute values on $|\Delta x_{\mu i}|$ and $|\Delta m_k|$ are to be taken element-wise.

A further simplification is possible, via the matrix determinant lemma, which states that for an invertible matrix $A$, and for vectors $u$ and $v$ there is equality between the following determinants

$$\det(A + u \cdot v^T) = (1 + v^T \cdot A^{-1} \cdot u)\det(A)$$

Although the equality is only valid for a single pair of vectors, it can be used as an approximation to calculate a perturbed matrix with many pairs of vectors. Fortunately, the $\Delta m_k$ parameters are supposed make the length of the $\frac{\partial F_\nu}{\partial m_k}\Delta m_k$ vectors small, compared to the matrix built from $\frac{\partial F_\nu}{\partial x_\mu}\Delta x_\mu$. To use the lemma, in the case of eq. 6 one of the vector pairs, $u_\nu$ is $\frac{\partial F_\nu}{\partial m_k}\Delta m_k$ with fixed $k$ index, that is summed over in the matrix product with the inverse of the $\frac{\partial F_\nu}{\partial x_\mu}$ matrix. The second of the pairs is not simply the unused $\Delta x_\mu$, but $\Delta x_\mu^{-1}$ and its element-wise product with the related $s(i, \mu, k)$ signs. The $s(i, \mu, k)$ signs don't need to be calculated though, since they were chosen to explicitly shrink the absolute value of the determinant of the Jacobian matrix $\frac{\partial F_\nu}{\partial x_\mu}$, and can be worked around using absolute values and a negative sign in the lemma. Hence the algorithm shall be, that one calculates the $v'_\mu = \left(\frac{\partial F_\nu}{\partial x_\mu}\right)^{-1} \cdot \frac{\partial F_\nu}{\partial m_k}|\Delta m_k|$ vector, takes the element-wise absolute values, and calculates the dot product of the resulting $|v'_\mu|$ vector by the $\frac{1}{|\Delta x_\mu|}$ vector. The result should be subtracted from unity, giving

$$V_{\Delta u_i} = \prod \Delta u_{\nu i} = \left(1 - \sum_k \left(\sum_\mu \frac{1}{|\Delta x_{\mu i}|}\left|\sum_\nu \left(\frac{\partial F_\nu}{\partial x_\mu}\right)^{-1} \cdot \frac{\partial F_\nu}{\partial m_k}|\Delta m_k|\right|\right)\right)\det\left(\frac{\partial F_\nu}{\partial x_\mu}\right) \qquad (7)$$

Here the index summations were explicitly written out. As eq. 6 and 7 are both approximations to the same order, one can expect similar results from both.

## 2.3 Optimized General Case

The general case in the previous subsection contains the $\Delta x_{\mu i}$ precision goal of each $x_{\mu i}$ data point explicitly. What values should one choose for them? One could go with a uniform value of $\Delta x$, or a non-isotropic $\Delta x_\mu$ for each dimension, but one can also expand the ideas in the subsection further. So, similar to using a constant $V_{\Delta u_i}$ volume with variable $\Delta u_{\nu i}$ instead of fixed $\Delta u_{\nu i}$ edges, one can sacrifice the precision goal of one $x_{\mu i}$ data element for an other's, still keeping the sum of $-\log V_{\text{data}} = -\log \Delta x_{\mu i}$ bits the same. As long as no $\log \Delta x_{\mu i}$ is set to zero, this is still a valid scheme, since the plan is to later take the limit of $\Delta x_{\mu i} \longrightarrow 0$. Hence, the method still incorporates every bit of $x_{\mu i}$, just renders certain data elements to be more important than others.

With this in mind, let's have a look into eq. 7. It isn't surprising from a linear approximation, the volume $V_{\Delta u_i}$ can be rewritten as a function of $r_{k\mu i} = \frac{\Delta m_k}{\Delta x_{\mu i}}$ ratios. Similarly, after a subtraction of right amounts of $\log(\Delta x_{\mu i})$s, the $\log(\Delta m_k)$ terms in $Q$ become $\log(r_{k\mu i})$. Though the maximization of the volume $V_{\Delta u_i}$ requires the variation of $\Delta x_{\mu i}$, but only with the constraints by either the local product $\prod_\mu \Delta x_{\mu i}$ being fixed, or by the more general means of keeping the $\prod_{i\mu} \Delta x_{\mu i}$ global product fixed. These constraints both allow the subtraction of $\log(\Delta x_{\mu i})$ terms from $Q$, without changing the position of the minima. Therefore $Q$ might be safely rewritten as a function of $r_{k\mu i} = \frac{\Delta m_k}{\Delta x_{\mu i}}$, though these variables are not independent.

It can also be seen, that now the likelihood $\left|\det\left(\frac{\partial F_\nu}{\partial x_\mu}\right)\right|$ is decoupled from the perturbations, which are supposed to be within the range $[0...1]$. It won't be used in this article, but it is interesting to see, that a different approximation may decouple each of the $\frac{\partial F_\nu}{\partial m_k}\Delta m_k$ terms further, giving the product

$$V_{\Delta u_i} = \prod \Delta u_{\nu i} = \left|\det\left(\frac{\partial F_\nu}{\partial x_\mu}\right)\right| \prod_k \left(1 - \sum_\mu \frac{1}{\Delta x_{\mu i}} \left|\sum_\nu \left(\frac{\partial F_\nu}{\partial x_\mu}\right)^{-1} \cdot \frac{\partial F_\nu}{\partial m_k}\Delta m_k\right|\right) \tag{8}$$

The substitution of $\log(V_{\Delta u}(x_\mu, m_k))$ into Q now results explicitly in a term containing the likelihood, while the addition of the logarithms of the perturbation terms act as a regularization upon the maximum likelihood criterium. The peculiarity of this approximation is, that now the terms containing the $\Delta m_k$ precision of a single parameter $m_k$ can be grouped together, including the explicit $-\log \Delta m_k$ term from $Q$ and the $x_{\mu i}$ datapoint-related terms from eq. 8. With a single $\Delta m_k$ parameter in each group, the optimization of $Q$ is decoupled, and the $\Delta m_k$ terms can be parallel estimated (still depending on $m_k$ and the explicit $\Delta x_{\mu i}$ though). For each $\Delta m_k$ the structure is simple, the additional parameters can be thought of as constants $a_i$ (depends on $k$, but that is omitted for the simplicity), and looks like $Q_k = -\log \Delta m_k - \sum_i \log(1 - a_i \Delta m_k)$

Nevertheless, the formulation of $V_{\Delta u_i}$ in eq. 7 allows a different simplification. As it can be seen, the summation over $\mu$ with the $\frac{1}{\Delta x_{\mu i}}$ factor can be moved one parenthesis further out, making it a dot product with the $\sum_k \left|\sum_\nu \left(\frac{\partial F_\nu}{\partial x_\mu}\right)^{-1} \cdot \frac{\partial F_\nu}{\partial m_k}\Delta m_k\right|$ vector. This means, that apart from taking the absolute values, it satisfies the criteria for the matrix determinant lemma, and in certain parameterizations eq. 7 can be an exact formula, not just an approximation. Furthermore, the formulation allows finding the optimum by varying the $\Delta x_{\mu i}$ terms with certain constraints, particularly the one that keeps the $\Delta x_{\mu i}$ volumes fixed, as $\prod_\mu \Delta x_{\mu i} = V_{\Delta x_i}$ for each data point $i$. The maximum value of the $\prod_\nu u_{\nu i} = V_{\Delta u_i}$ volume is then taken at

$$\Delta V_{u_i}^{\max} = \left|\det\left(\frac{\partial F_\nu}{\partial x_\mu}\right)\right| \left(1 - n_{\dim} \frac{\sqrt[n_{\dim}]{\prod_\mu \left(\sum_k \left|\sum_\nu \left(\frac{\partial F_\nu}{\partial x_\mu}\right)^{-1} \cdot \frac{\partial F_\nu}{\partial m_k}\Delta m_k\right|\right)}}{\sqrt[n_{\dim}]{V_{\Delta x_i}}}\right)$$

Substituting these terms into $Q$, gives the log-likelihood and a sum of logarithms of the correction terms, in the form of $-\sum_i \log\left(1 - a_i/\sqrt[n_{\dim}]{V_{\Delta x_i}}\right)$. Its minimum value with respect to $V_{\Delta x_i}$, with the constraint $\prod_i V_{\Delta x_i} = V_{\text{data}}$ can be expressed in a closed formula as

$$\min_{V_{\Delta x_i}}\left(-\sum_i \log(\Delta V_{u_i}^{\max})\right) = -n_{\text{data}}\log\left(1 - \sqrt[n_{\text{data}}]{\dim\sqrt{\frac{\prod_{i\mu}\sqrt[n_{\dim}]{\sum_k \left|\sum_\nu \left(\frac{\partial F_\nu}{\partial x_\mu}\right)^{-1} \cdot \frac{\partial F_\nu}{\partial m_k}\Delta m_k\right|}}{V_{\text{data}}}}}\right) \quad (9)$$

Here, signature of the $\Delta m_k$ terms is the opposite as of the $-\log\Delta m_k$ terms, the bit length of the $m_k$ parameters in $Q$. The consequence is, that during the minimization of $Q$, large $\Delta m_k$ will be favored due to the $-\log\Delta m_k$ terms, and small $\sum_\nu \left(\frac{\partial F_\nu}{\partial x_\mu}\right)^{-1} \cdot \frac{\partial F_\nu}{\partial m_k}$ terms relative to $\Delta m_k$, due to the terms in eq. 9. Heuristically, we may think of the $\left(\frac{\partial F_\nu}{\partial x_\mu}\right)^{-1}$ term as the inverse of the likelihood, which has its own term in $Q$ and maximized, so the effects of $\left(\frac{\partial F_\nu}{\partial x_\mu}\right)^{-1}$ tends to be small. As this is a matrix, either its elements are small or the following $\frac{\partial F_\nu}{\partial m_k}\Delta m_k$ vectors are parallel to the eigenvectors with the smallest eigenvalue. With similar reasoning, for fixed $m_k$ parameters we can treat $\Delta m_k$ as a vector that has a dot product with the multitude of $v_k^{(i\mu)} = \left|\sum_\nu \left(\frac{\partial F_\nu}{\partial x_\mu}\right)^{-1} \cdot \frac{\partial F_\nu}{\partial m_k}\right|$ vectors. The normalization of the $\Delta m_k$ vector isn't built into the equations, but as $\prod_k \Delta m_k$ is nonzero, the length is nonzero as well. Adding these up, we can expect the $Q$-minimizing $\Delta m_k$ vector to be perpendicular to most of the $v_k^{(i\mu)}$ vectors, near the direction of the principal axis[6] with the smallest eigenvalue of the $V_{k,l} = \sum_{i\mu} v_k^{(i\mu)} v_l^{(i\mu)}$ matrix.

The restrictions to eq. 7 and 8 when applied in $Q$ are simple. One constraint is, that the $\Delta m_k$ and $\Delta x_{\mu i}$ terms need to be smaller or equal to one, as a negative bit length contribution is regarded as unrealistic. The second one is, that each individual perturbation term, the ones summed up by $k$ in eq. 7 and the ones in the product over $k$ in eq. 8, should not be equal to or larger than one. This requirement is due to the inequality 4, that was later formulated as the constraint, that the $\frac{\partial F_\nu}{\partial m_k}\Delta m_k$ terms can only reduce the $\left|\det(\frac{\partial F_\nu}{\partial x_\mu})\prod \Delta x_\mu\right|$ volume up to zero. And zero volume itself is not allowed when a logarithm is applied to it in $Q$. An indirect restriction applies to the $m_k$ parameter set, as the parameters must be finite. The method uses the $\Delta m_k$ auxiliary variables to track the length of the $m_k$ parameters, but only after the decimal point. This is a valid approximation, as long as we are interested in the $\Delta m_k \to 0$ limit, and the bits before the decimal point can be neglected. This assumption would break down if a parameter was infinite, as there would be infinitely many bits before the decimal point.

## 2.4 In Essence

As can be seen, this method is optimized for on-demand data refinement. It can reconstruct the original data $x_{\mu i}$ with a desired precision $\Delta x_{\mu i}$, and when more bits of $x_{\mu i}$ needed, one only needs corrections within the calculated $\Delta u_{\nu i}$ and $\Delta m_k$ range. With the requirement of minimizing $Q$, we can be sure that we only need minimal number of bits to deal with for the $x_{\mu i}$ correction. Thus the method is the optimal way either to literally describe data, or for streaming the data in communication channels. This gave the title of the article, Occam's Ghost, as the method is designed to compress information of transcendental numbers. However, the most important aspect of the method is that the minimization of $Q$ gives a weak ordering to the bits of the model parameters, and also to the data points and dimensions via the variation of $\Delta x_{\mu i}$. The ordering then gives us the most important bits needed to reconstruct the dataset with a given overall precision. It can be thought of as an series expansion along the most important bits of the data. Still, many situations won't allow the reweighing of the $\Delta x_\mu$, e.g. proper probability density estimations. Even in those cases, the aforementioned optimized case may help, as the optima can be approximated more easily, and that can be used as a starting point for further optimizations.

# 3 Recipe for the Impatient

## 3.1 From Probabilities to Coordinates

Usually one does not start with a parameterized coordinate system $F_\nu(x_\mu, m_k)$, but tries to find a suitable coordinate system for a parametric probability density function, $f(x_\mu, m_k)$. Building a coordinate system is rather complicated, as one generally wants one that has only a couple of coordinate singularities, where $F_\nu(x_\mu)$ is a so-called conservative vector field, so path integrals depend only on the endpoints of the paths. Such is not easily constructed with constraints like $\left|\det\left(\frac{\partial F_\nu}{\partial x_\mu}\right)\right| = f(x_\mu)$. Fortunately, a PDF can be factorized into a product of marginal and conditional distributions as

$$f(x_\mu) = \text{Prob}(x_1)\text{Prob}(x_2|x_1)\text{Prob}(x_3|x_1, x_2)...$$

This is a product of terms with restricted dependence on the $x_\mu$ input coordinates. One can define the model coordinates as

$$F_1(x_1) = \int_{-\infty}^{x_1} \text{Prob}(x_1')\text{dx}_1'$$

$$F_2(x_2, x_1) = \int_{-\infty}^{x_2} \text{Prob}(x_2'|x_1)\text{dx}_2'$$

$$F_3(x_3, x_2, x_1) = \int_{-\infty}^{x_3} \text{Prob}(x_3'|x_2, x_1)\text{dx}_3'....$$

and so on. Their corresponding $x_\mu$ derivatives create a triangular matrix, $\frac{\partial F_\nu}{\partial x_\mu}$, thus its determinant is the desired product of conditional distributions, giving $f(x_\mu)$. This also means that the off-diagonal elements won't contribute to the end result, the determinant related to the likelihood, but are still necessary to calculate the perturbed determinant via eq. 7 or 8.

Although the method restricts the available mappings of the data, it is often enough, when one is only interested in a probability density estimate.

## 3.2 The Unitary Constraint of Amplitudes

Most models are a sum of different individuals PDFs weighted with an $a_j \in [0...1]$ amplitude, which need to be normalized as $\sum_j a_j = 1$. Equality constraints would introduce an infinitely precise fine tuning to the $\Delta a_j$ volumes where their rounding is may happen. Technically it is possible remove that infinity from $Q$, or remove the extra degree of freedom, but that seems to be a complicated step. Instead, re-parameterizations like

$$a_j' = \frac{a_j^2}{\sum_l a_l^2}$$

are allowed. It is easy to implement it into the derivatives of $F_\nu$ via the chain rule, and has the advantage that the derivates are continuous around $a_j'=0$, aiding the minimization toward small amplitudes. In this parameterization the amplitudes can be independently varied, hence a $-\log\Delta a_j$ term for them in $Q$ properly measures their bit-requirement.

A different approach is a bit more complex, but properly addresses the degrees of freedom. One needs to take angles, $\alpha_j = [0...2\pi)$, where $j = 1, 2, ...., n_a - 1$, one fewer than $n_a$, being the number of PDFs. With such angles, the rotation of a $e = (0, 0, ..., 0, 1)$ unit vector may take the form of $e_i = \sin(\alpha_i) \prod_{j<i} \cos(\alpha_j)$ for $i \in [1, ..., n_a - 1]$ and $e_{n_a} = \prod_{j<n_a} \cos(\alpha_j)$ for the last component. As the vector is square-normalized, the square of each component can be used as a probability amplitude for PDF summation. Though the formula is not symmetric, it and its derivatives are rather simple, and easy to implement.

## 3.3 Regularized Fit of Probability Density Functions

The takeaway point of the article is, that one can regularize probability density function fitting, or maximum likelihood fitting by following a simple recipe. The method is based on compressing the $x_{\mu i}$ data points, truncated after $-\log \Delta x_{\mu i}$ bits, with a model $u_{\nu i} = F_\nu(x_{\mu i}, m_k)$ using only $Q = -\sum_{\nu_i} \log(\Delta u_{\nu i}) - \sum_k \log(\Delta m_k)$ number of bits. To gain more bits on compression, it is possible vary the model parameters $m_k$, the auxilary model parameters $\Delta m_k$ and the compression targets. The latter targets are the $-\log \Delta x_{\mu i}$ number of bits, varied by either keeping $-\sum_\mu \log \Delta x_{\mu i}$ constant for each $i$ data point (local variation), or keeping only the sum of all bits of all of the data points, $-\sum_{\mu i} \log \Delta x_{\mu i}$ constant (global variation).

A part of the loss function $Q$ that need to be minimized is the log-likelihood

$$Q_l = -\sum_i \log(f(x_{\mu i}, m_k)) = -\sum \log\left(\left|\det\left(\frac{\partial F_\nu(x_{\lambda i}, m_k)}{\partial x_\mu}\right)\right|\right).$$

As it can be handled separated from the other parts, the determinant calculation is not necessary when the PDFs are readily available. However, the $\frac{\partial F_\nu(x_{\lambda i}, m_k)}{\partial x_\mu}$ matrix, or its inverse is still needed for the regularization part, but it is usually related to some of the parametric derivatives (e.g. in a Gaussian mixture model, if $m_{\mu j}$ are the means if the individual Gaussians, $\frac{\partial F_\nu(x_{\lambda i}, m_k)}{\partial x_\mu} = -\sum_j \frac{\partial F_\nu(x_{\lambda i}, m_k)}{\partial m_{\mu j}}$).

The method with the fewest parameters is which lets vary the $\Delta x_{\mu i}$ precision targets within a fixed $-\sum_{\mu i} \log \Delta x_{\mu i} = -\log(V_{\Delta x})$ overall bit-count. That only requires a single additional $\Delta m_k$ helper variable for each $m_k$ parameter of the model. In principle $\Delta m_k$ has dependence on the $\Delta x_{\mu i}$ after minimizing $Q$, but the $\frac{\Delta m_k}{\Delta x_{\mu i}}$ ratios won't change. Hence, if one is only interested in the $\Delta x \to \infty$ limit, it is enough to treat $\Delta m_k$ as a ratio of $\frac{\Delta m_k}{\sqrt[n_{\text{data}}]{V_{\Delta x}}}$ itself, and pretend $V_{\Delta x} = 1$. Their direct contribution to $Q$ is their bit length (they represent the truncation length of $m_k$)

$$Q_\Delta = -\sum_k \log(\Delta m_k)$$

The current method is valid for $\Delta m_k \in [1...0)$, but there is no known theoretical reason not to expand it later for $\Delta m_k > 1$.

The last part may change, depending on the constraints of $\Delta x_{\mu i}$, but the one incorporating the $-\sum_{\mu i} \log \Delta x_{\mu i} = 1$ is the following

$$Q_r = -n_{\text{data}} \log\left(1 - n_{\text{dim}} \sqrt[n_{\text{data}} \cdot n_{\text{dim}}]{\prod_{\mu i}\left(\sum_k \left|\sum_\nu \left(\frac{\partial F_\nu(x_\lambda, m_l)}{\partial x_\mu}\right)^{-1} \cdot \frac{\partial F_\nu(x_\lambda, m_l)}{\partial m_k} \Delta m_k\right|\right)}\right)$$

As the term contains a large number of multiplications and takes its $n_{\text{data}} \cdot n_{\text{dim}}$-th root, it is advised to use the sum of the logarithms and dividing it by the number of data points and the length of the data point vectors. In case the part within the logarithm becomes smaller than zero, it should be regarded as invalid (or the $Q_r$ as positive infinite), and at least one of the $\Delta m_k$ terms need to be decreased to make it finite again. The sum of the three terms is the overall bit count of the model, the target of minimization $Q = Q_l + Q_\Delta + Q_r$.

# 4　Error Estimation

A simple estimate of the $m_k$ parameter variance from changing the underlying data requires only that the model $F(x, p_k)$ is sufficiently smooth and its parameters are minimizing a loss function $Q(x_{\mu i}, p_k)$, so $\frac{\partial Q}{\partial p_k} = 0$. (Remark: In this section, the $p_k$ parameters refer to the full parameter set, $m_k$, concatenated with the $\Delta m_k$ truncation volumes, and $\Delta p_k$ refers actual to deviations from this parameter set, due to variations in the data.) The estimation is modeled with a Gaussian perturbation in the $u_{\nu i} = F_\nu(x_{\mu i}, p_k)$ image space, propagated to $x_{\mu i}$ and $p_k$. The perturbation of the $F_{\nu i}$ map is by definition

$$\Delta F_{\nu i} = \frac{\partial F_{\nu i}}{\partial x_\mu} \Delta x_{i\mu} + \frac{\partial F_{\nu i}}{\partial p_k} \Delta p_k = \frac{\partial F_{\nu i}}{\partial x_\mu} \left(\frac{\partial F_{\lambda i}}{\partial x_\mu}\right)^{-1} \left(\Delta u_{\lambda i} - \frac{\partial F_{\lambda i}}{\partial p_l} \Delta p_l\right) + \frac{\partial F_{\nu i}}{\partial p_k} \Delta p_k = \Delta u_{\nu i}$$

It is a distortion on $F_{\nu i}$ and on the likelihood $\left|\det\left(\frac{\partial F_{\nu i}}{\partial x_{\mu i}}\right)\right|$.

From the polynomial expansion of the loss function around $x^0_{\mu i}$ and $p^0_k$,

$$\begin{aligned} Q(x^0_{\mu i} + \Delta x_{\mu i}, p^0_k + \Delta p_k) &= Q(x^0, p^0) + \sum_i \frac{\partial Q(x^0, p^0)}{\partial x_{\mu i}} \Delta x_{\mu i} + \\ &+ \frac{\partial Q(x^0, p^0)}{\partial p} \Delta p + \frac{1}{2} \frac{\partial^2 Q(x^0, p^0)}{\partial p_k \partial p_l} \Delta p_k \Delta p_l + \\ &+ \sum_i \frac{\partial^2 Q(x^0, p^0)}{\partial p_k \partial x_{\mu i}} \Delta p_k \Delta x_{\mu i} + \sum_{i,j} \frac{\partial^2 Q(x^0, p^0)}{\partial x_{\nu j} \partial x_{\mu i}} \Delta x_{\nu j} \Delta x_{\mu i}. \end{aligned}$$

At optimum, the first $p_k$ derivative is zero, $\frac{\partial Q}{\partial p_k} = 0$, and so is the $\frac{\partial Q}{\partial \Delta p_k}$ at the optimum with the perturbed sample, $x_{\mu i} + \Delta x_{\mu i}$. This gives $\frac{\partial^2 Q(x^0, p^0)}{\partial p_k \partial p_l} \Delta p_l = -\sum_i \frac{\partial^2 Q(x^0, p^0)}{\partial p_k \partial x_{\mu i}} \Delta x_{\mu i}$, a connection of the sample's perturbation to the inferred optimal parameters. As long as the Hessian, $\frac{\partial^2 Q(x^0, p^0)}{\partial p_k \partial p_l}$, is invertible, the standard deviation of $\Delta p_k$ can be expressed with the error propagation formula. To get the estimate for the standard deviation of the $x_{\mu i}$ sample, we can turn to the fact that for a good model, the $u_{\nu i} = F(x_{\mu i}, p_k)$ encoded data is uniformly distributed, and take $\sigma_{\Delta u_{\nu i}} = (12 n_{\text{data}})^{-1/2}$ as the standard deviation of the encoded data point (scaling may depend on the boundary conditions). With this $\Delta u_{\nu i}$, $\Delta x_{\mu i} = \left(\frac{\partial F_{\nu i}}{\partial x_\mu}\right)^{-1} \Delta u_{\nu i} - \left(\frac{\partial F_{\nu i}}{\partial x_\mu}\right)^{-1} \left(\frac{\partial F_\nu(x_{\lambda i}, p_k)}{\partial p_l}\right) \Delta p_l$. For a first approximation, the second term might be neglected, assuming that the $\Delta p_l$ deviations have a small effect only. The more precise handling takes that into account as well, reflecting the fact that the estimate of the standard deviation of $x_{\mu i}$ is coming from a model. The end result is

$$\Delta p_l = -\left(\frac{\partial^2 Q(x^0, p^0)}{\partial p_k \partial p_l} - \sum_j \frac{\partial^2 Q(x^0, p^0)}{\partial p_k \partial x_{\alpha j}} \left(\frac{\partial F_{\beta j}}{\partial x_{\alpha j}}\right)^{-1} \frac{\partial F_{\beta j}(x^0_j, p^0)}{\partial p_l}\right)^{-1} \cdot \sum_i \frac{\partial^2 Q(x^0, p^0)}{\partial p_k \partial x_{\gamma i}} \left(\frac{\partial F_{\nu i}}{\partial x_{\gamma i}}\right)^{-1} \Delta u_{\nu i}.$$

With $\sigma_{\Delta u_{\nu i}} = (12 n_{\text{data}})^{-1/2}$, the standard deviation of the $p_k$ can be estimated via the formula for the propagation of uncertainty. The inverse of $\frac{\partial F_{\beta i}}{\partial x_{\alpha i}}$ is taken for each $i$ and $j$ index, while the summation of $\mu, \nu, \alpha, \beta$ and $\gamma$ are implicitly assumed.

So far, this expression was general, and could be applied to almost any loss function. However, it still requires the model $F$ to be a map from $x_{\mu i}$ to $u_{\nu i}$, as the uncertainty for the data points are estimated from the standard deviation of the unitary distribution. For the method described in this article, the parameters $p_k$ include the $m_l$ parameters of $F_\nu$, and also the auxiliary parameters $\Delta m_l$, from which $F_\nu$ has no dependence. $Q$ in our case contains the likelihood, hence $\frac{\partial^2 Q(x^0, p^0)}{\partial p_k \partial p_l}$ contains the Fisher information[16] as well (for the $m_k$ parameters within $p_k$). Apart from the possibility, that $\sigma_{p_l}$ can be made small when $\frac{\partial^2 Q(x^0, p^0)}{\partial p_k \partial p_l}$ is large and $\sum_i \frac{\partial^2 Q(x^0, p^0)}{\partial p_k \partial x_{\gamma i}}$ is small, there is no apparent advantage of one method or an other. Optimization procedures usually look for a global optimum of $Q$, disregarding its second derivatives, but it nevertheless expresses the uncertainty of the parameters.

## 5 Properties of the Univariate Case

Analysis of the univariate case can reveal some of the properties of the method. Writing down the objective function for the $r_k = \frac{\Delta m_k}{\Delta x}$ ratios gives:

$$Q(m_k, r_k) = -\sum_i \log\left( f(x_i, m_l) - \sum_k \left|\frac{\partial F(x_i, m_l)}{\partial m_k}\right| r_k \right) - \sum_k \log(r_k) \tag{10}$$

Where $f(x_i, m_l) = \left|\frac{\partial F(x_i, m_l)}{\partial x}\right|$ is the univariate PDF equivalent and $u_i = F(x_i, m_k)$ is a cumulative distribution function. Without the $r_k$ parts the minimum of $Q$ gives the parameterization with maximum likelihood. We can note that due to the last term in the eq. 10, the $r_k$ terms need to be greater than zero, meaning that as long as a $\frac{\partial F}{\partial m}$ term is not zero, they contribute to the first term. In the form $Q = -\sum_i \log(f_i) - \sum_i \log\left(1 - \sum_k \frac{\partial F}{\partial m_k} r_k / f_i\right) - \sum_k \log(r_k)$, it is visible, that the $r_k$ terms are somewhat independent from the likelihood-related terms. A single $r_k$ have the ability to affect multiple $f(x_i, m_l)$ terms during the minimization.

Similar to likelihood maximization, minimal $Q$ favors large $f(x_i, m_l)$ values, but unlike the classical method it has a preference for smaller $\left|\frac{\partial F}{\partial m_k}\right| r_k$. Furthermore, it prefers small $\left|\frac{\partial F}{\partial m_k}\right|$ while allows and prefers large $r_k$, due to the ineq. 4. On the similar train of thought, if a given PDF can be represented by two different parameter sets, $m_{1l}$ and $m_{2l}$, and $f(x_i, m_{1l}) = f(x_i, m_{2l})$ which only differ by some of the derivatives $\frac{\partial F}{\partial m}$, minimizing $Q$ may favor one of those. For example, in case the derivatives are similar for $i > 3$, $\left|\frac{\partial F(x_i, m_{1l})}{\partial m_i}\right| = \left|\frac{\partial F(x_i, m_{2l})}{\partial m_i}\right|$, but differ for $i = 1$ as $\left|\frac{\partial F(x_i, m_{1l})}{\partial m_1}\right| = 0 \neq \left|\frac{\partial F(x_i, m_{2l})}{\partial m_1}\right|$ and for $i = 2, i = 3$ as $\left|\frac{\partial F(x_i, m_{1l})}{\partial m_2}\right| \neq 0 = \left|\frac{\partial F(x_i, m_{2l})}{\partial m_2}\right|$ and $\left|\frac{\partial F(x_i, m_{1l})}{\partial m_3}\right| \neq 0 = \left|\frac{\partial F(x_i, m_{2l})}{\partial m_3}\right|$, $Q$ would favor the $m_{2l}$ parameter set, the one with more small derivatives via allowing $r_2$ and $r_3$ to be large, and contributing less to the $-\log(r_1) - \log(r_2) - \log(r_3)$ part than in the $m_{1l}$ case.

- Thus, minimizing $Q$ favors models with fewer parameters, or parameters with vanishing derivatives. This also means, that in mixture models where the final CDF is built by summing up individual CDFs with parametric amplitudes are expected to favor fewer components, while the obsolete ones converge to zero.

Minimizing $Q$ gives the model parameters $m_k$ and $u_{\nu i}$ where a $\Delta m_k$ and $\Delta u_{\nu i}$ error would result in an $\Delta x_{\mu i}$ change at maximum, but that relation can be reversed. A $\Delta x$ change of a data point $x_{\mu i}$ could be compensated with a change in the parameterization by $\Delta m_k$ and $\Delta u_{\nu i}$ at maximum, due to the total derivative used for eq. 3, though it might not be a minimum of $Q$ on the new dataset. However

- Replacing the complete dataset with something from the same distribution while minimizing $Q$ will result in tractable changes in the parameterization. See Section 4 on error estimation.

Although the method gives and maximizes a combined upper bound for the change in $m_k$ and $u_{\nu i}$, that should be compared to the no bounds given by an unregularized likelihood maximization. A bare maximum likelihood approach may lead to phase spaces with much larger $\frac{\partial F_i}{\partial m_k}$ derivatives, as it only focuses on $\left|\frac{\partial F_i}{\partial x}\right|$.

The method was designed with the property that assures, no perturbation would happen to $m_k$ and $u_{\nu i}$ larger than $\Delta m_k$ and $\Delta u_{\nu i}$ when reconstruction is required smaller than $\Delta x_{\mu i}$. The minimization of $Q$ also provides that these perturbations can be done with minimal number of bits.

- Thus, the method is ideal for streaming the model, as it gives the most important bits required to reconstruct the dataset with increasing precision. Furthermore, this allows us to be imprecise, tells us which parameters could be neglected.

The method prefers small $\left|\frac{\partial F}{\partial m_k}\right| r_k$, but due to the $-\log(r_k)$ term in $Q$, it prefers logarithmically large $r_k$, and therefore parameterization with exponentially small $\left|\frac{\partial F}{\partial m_k}\right|$.

- Mixture models, with parametric probability amplitudes that might be set to zero, allow the $\left|\frac{\partial F}{\partial m_k}\right|$ derivatives to be small, hence a component of $F$ that depends on $m_k$ can converge to zero.

The requirement that $Q$ is minimized favors large $\Delta m$ in some respect, that means large allowed change in the model parameters. This is achieved by $m$ values where the $\frac{\partial F}{\partial m}$ derivatives are small, and the parameters with significantly large $\Delta m$ range can be mostly neglected. The consequence of fewer parameters then usually means smaller standard deviation of the remaining $m$ parameters, though those are not necessarily the same order of magnitude as their $\Delta m$.

The indirect effect of fewer parameters is then smoother $\frac{\partial F}{\partial x}$, and generally small spatial derivatives of the modeling PDF, $\frac{\partial^2 F}{\partial x^2}$. This is because in $Q$ the $\log(\Delta u)$ term requires $\Delta u$ to be positive around a data point, and a large $\frac{\partial^2 F}{\partial x^2}$ means that $f$ has a peak around a cluster of data points. However, the aforementioned fewer parameters typically mean that fewer clusters can be described, thus it altogether means smoother PDFs. Nevertheless, there could be cases, when there are hidden symmetries in the data points, e.q. they lay on an under-sampled grid. In that case, a Fourier model could reveal the repetitive nature of the data, by favoring some high frequency component which would mean a highly non-smooth, but still simple (low parameter number) model.

The minima of Q is not necessarily a measure of fitness, as that value is only useful with the combination of a $\Delta x$. Furthermore, constraints like $\Delta m$ being smaller than one are not built into the method, and parameters that have no effect on the model could have diverging $\Delta m$, giving arbitrarily small $Q$. The method should be used more like a decision function, where two models are compared not by the value of $Q$, but by putting the combination of the two models into $Q$, and letting the method reject one or the other.

## 6 Implementation of Gaussian Mixture Model

To demonstrate the feasibility of the, the algorithm explained in Sections 2.2, 2.3 and 3, including global and local $\Delta x_{\mu i}$ variation was implemented[10] for gaussian mixture model in the Julia[2] language. Julia was chosen mainly for the wide variety of available libraries, especially the field of automatic differentiation and numerical optimization, that allowed fast development and testing.

For the sake of simplicity, the covariance matrices of the individual Gaussians were restricted to the diagonal terms. The modeling function was based on the integrated marginal probability distributions, as mentioned in Section 3.

For $n_a$ number of $a_i$ amplitudes, $m_{\mu i}$ means and $\sigma^2_{\mu i}$ standard deviations the probability density function wished to be modeled is

$$\text{Prob}(x_\mu) = \sum_i a_i \frac{1}{\sqrt{2\pi \prod_\nu \sigma^2_{i\nu}}} \exp\left(-\prod_\nu \frac{(x_\nu - m_{i\nu})^2}{2\sigma^2_{i\nu}}\right)$$

The marginal distributions needed are

$$p_\lambda(x_1, x_2, ..., x_\lambda) = \sum_i a_i \frac{1}{\sqrt{2\pi \prod_{\nu \leqslant \lambda} \sigma^2_{i\nu}}} \exp\left(-\prod_{\nu \leqslant \lambda} \frac{(x_\nu - m_{i\nu})^2}{2\sigma_{i\nu}^2}\right),$$

from which the conditional distributions are relatively simple

$$\text{Prob}(x_\lambda | x_{\lambda-1}, ...., x_1) = \frac{\text{Prob}(x_\lambda, x_{\lambda-1}, ...., x_1)}{\text{Prob}(x_{\lambda-1}, ...., x_1)} = \frac{p_\lambda(x_\mu)}{p_{\lambda-1}(x_\mu)}$$

The marginal cumulative density functions producing the above PDFs when derived are

$$\text{CDF}_\lambda(x_\mu) = \int_{-\infty}^{x_\lambda} \frac{p_\lambda(x_\mu)}{p_{\lambda-1}(x_\mu)} = \frac{\sum_i a_i \text{erf}\left(\frac{x_\lambda - m_{i\lambda}}{|\sigma_{i\lambda}|}\right) \prod_{\mu < \lambda} \exp\left(-\frac{(x_\nu - m_{i\nu})^2}{2\sigma^2_{i\nu}}\right) \frac{1}{\sqrt{2\pi\sigma^2_{i\nu}}}}{\sum_i a_i \prod_{\mu < \lambda} \exp\left(-\frac{(x_\nu - m_{i\nu})^2}{2\sigma^2_{i\nu}}\right) \frac{1}{\sqrt{2\pi\sigma^2_{i\nu}}}}$$

Where $\text{erf}(x)$ is the error function. These CDFs are ready to be used as a $F_\lambda(x_\mu) = \text{CDF}_\lambda(x_\mu)$ coordinate system for modeling the data.

It must be noted, that the use of the integral of the Gaussian allows the introduction of a new free parameter, depending on the start of the integration, an integral constant. Though the Jacobi determinant of the CDF won't contain this constant, the Jacobian matrix and the parametric derivatives still contain it. Most forms of the method in this article contain the product of the derivative and the truncation range of the probability amplitude $\frac{\partial F_\nu}{\partial a_i}\Delta a_i$, as in eq. 9, where the integration constant directly affects the allowed magnitude of $\Delta a_i$. Terms similar to $-\log\left(f(x) - \left|\frac{\partial F_\nu}{\partial a_i}\right|\Delta a_i\right)$ appear in the $Q$ function, subject to minimization, where the erf($x$) error function and the integral constant directly appears. For $x \to \pm\infty$ the $f(x)$ probability density function is small, but the error function is dominated by the value of the integral constant. Those regions thus have an effect on the maximal $\Delta a_i$ value, the truncation region of the probability amplitude $a_i$, and also on the minimal $f(x)$ probability density, the likelihood of an $x_\mu$ data point in the tail region.

## 6.1 Structure

The sample code is able to perform fitting a PDF of mixture of Gaussians with diagonal covariance matrices in arbitrary dimensions. By default, a sample with three slightly overlapping Gaussian is generated in two dimensions, one large component, with 80% probability and two smaller, with 10% each. A univariate sample is later generated by projecting the sample to one of the axes.

The core of the method calculates the parametric gradient of the marginal probability based coordinates, and passes it to the $Q$ bit requirement calculation. The gradient of $Q$ is only calculated numerically, for those minimization algorithms that require it.

For sample points, which lie far from all the Gaussian centers by many standard deviation, certain calculations are simplified. Taking the logarithm of the sum of exponentials would be numerically false, and it is approximated by expansion around on the largest term. This avoids many of the $\log(0)$ pitfalls and helps fitting the tails of the model.

Optimization is performed using a chain of three gradient free methods, and a gradient based one at the end, all from the NLopt optimization package[8]. In order, they are a simplex based method (Sbplex)[14] starting from a random point, a genetic algorithm (ESCH)[1][13][4][3] performing a global search, then another simplex method (Sbplex) refining the current best value, and finally a gradient based method (MMA, Method of Moving Asymptotes)[15].

## 6.2 Performance

Fitting a small sample from a mixture of three Gaussians typically results in one or two significant peaks, while larger samples max out at three components. An example is shown on fig. 1. Exceptions are, when a data point lies far from other significant peaks, and those are classified as outliers, peaks on their own.

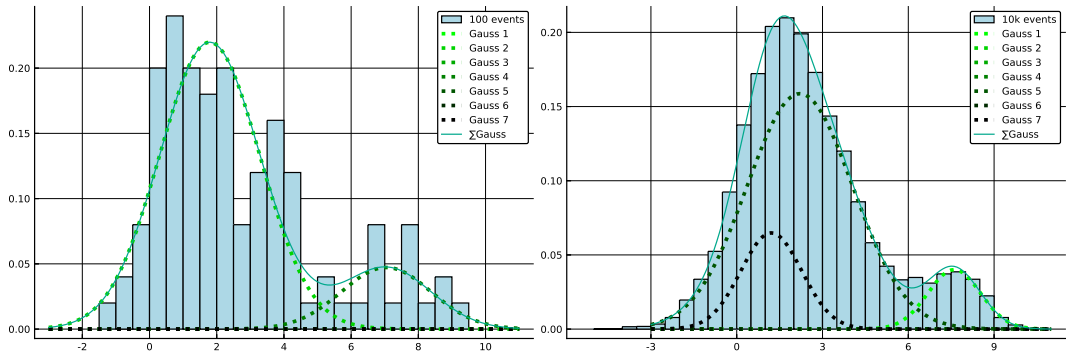

**Figure 1.** The left figure shows a run on 100 events, fitted best with two Gaussians, while the figure on the left was produced on 10k events, having three significant Gaussians.

Using different samples from the same distribution, one can expect similar fitting results. As fig. 2. shows, some samples have a pronounced secondary peak, while others prefer a single peak. Nevertheless, the $Q$ values of the given model on the other samples are close to each other. As each sample contain a 100 events, on can expect deviations on the order of $\sqrt{100}$ percent, the Poisson error coming from bootstrapping the sample and calculating $Q$. The deviations, when normalized to the $Q$ value of training sample can be seen in table 1., and they are close the approximate Poisson error.

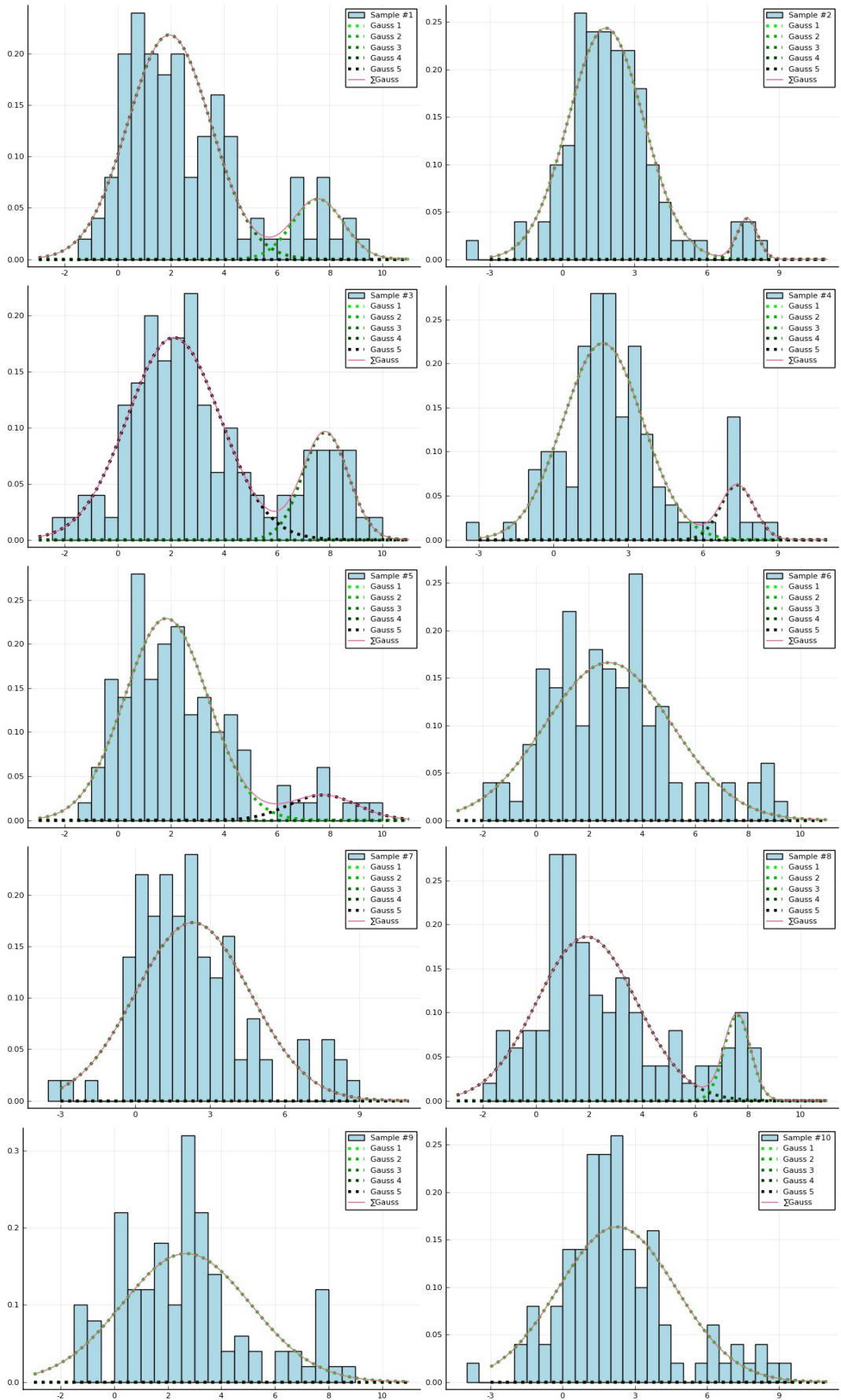

**Figure 2.** The figures show the optimal Gaussian composition for ten different samples from the same distribution. The fitting procedure allowed a maximum of five Gaussians, but the optima contains only one or two significant peaks.

|  | testing sample | | | | | | | | |
|---|---|---|---|---|---|---|---|---|---|
| 0.0% | -5.5% | 7.0% | -2.4% | -0.9% | 4.1% | -0.3% | 3.1% | 0.5% | 1.8% |
| 18.5% | 0.0% | 33.7% | 9.4% | 15.7% | 22.2% | 11.1% | 16.9% | 14.0% | 19.4% |
| -3.9% | -7.0% | 0.0% | -5.3% | -3.6% | -0.9% | -3.4% | -1.1% | -3.5% | -1.3% |
| 4.6% | -3.8% | 14.4% | 0.0% | 4.5% | 8.9% | 2.3% | 6.5% | 3.6% | 6.3% |
| 3.1% | -4.9% | 12.2% | 0.4% | 0.0% | 5.8% | 1.5% | 6.8% | 3.7% | 3.3% |
| 1.5% | -5.0% | 9.6% | -2.0% | -0.6% | 0.0% | -1.1% | 3.4% | -0.1% | 1.4% |
| 4.2% | -5.2% | 15.1% | -0.2% | 0.6% | 2.6% | 0.0% | 5.9% | 2.3% | 2.6% |
| 0.8% | -7.6% | 8.6% | -3.9% | 1.0% | 4.6% | -2.4% | 0.0% | -1.4% | 2.4% |
| 1.6% | -5.1% | 10.0% | -1.9% | -0.6% | 0.2% | -1.1% | 3.6% | 0.0% | 1.5% |
| 1.5% | -6.8% | 11.2% | -2.4% | -1.7% | 0.1% | -2.3% | 3.0% | -0.1% | 0.0% |

(rows labeled "training sample")

**Table 1.** Relative $Q$ values of the optimal models on the different 1D samples. Most deviations are within 10%, as it can be expected from a variation of 100 data points on a smooth PDF.

|  | testing sample | | | | | | | | |
|---|---|---|---|---|---|---|---|---|---|
| 0.0% | -5.5% | 7.4% | -2.4% | -0.8% | 4.5% | 0.0% | 3.4% | 0.7% | 2.2% |
| 16.4% | 0.0% | 31.5% | 8.5% | 15.4% | 21.3% | 10.0% | 16.0% | 12.9% | 17.9% |
| -4.0% | -7.1% | 0.0% | -5.5% | -3.6% | -0.7% | -3.3% | -1.0% | -3.6% | -1.1% |
| 4.5% | -3.7% | 14.5% | 0.0% | 4.6% | 9.3% | 2.5% | 6.7% | 3.6% | 6.6% |
| 3.1% | -4.9% | 12.6% | 0.4% | 0.0% | 6.1% | 1.8% | 7.1% | 3.9% | 3.6% |
| 1.5% | -5.2% | 9.9% | -2.1% | -0.7% | 0.0% | -1.1% | 3.5% | -0.1% | 1.5% |
| 4.3% | -5.3% | 15.7% | -0.2% | 0.7% | 2.7% | 0.0% | 6.1% | 2.4% | 2.8% |
| 0.7% | -7.6% | 8.8% | -4.1% | 1.4% | 5.0% | -2.4% | 0.0% | -1.4% | 2.7% |
| 1.7% | -5.3% | 10.3% | -2.0% | -0.7% | 0.1% | -1.1% | 3.7% | 0.0% | 1.5% |
| 1.6% | -7.1% | 11.6% | -2.5% | -1.8% | 0.1% | -2.4% | 3.1% | -0.1% | 0.0% |

**Table 2.** Relative entropy of the optimal models on the different 1D samples.

|  | testing sample | | | | | | | | |
|---|---|---|---|---|---|---|---|---|---|
| 0.0% | -0.1% | 2.0% | 2.7% | -4.0% | -1.1% | 3.5% | ∞ | ∞ | ∞ |
| ∞ | 0.0% | ∞ | ∞ | ∞ | ∞ | ∞ | ∞ | ∞ | ∞ |
| 1.4% | 4.5% | 0.0% | 6.6% | -4.0% | -0.2% | ∞ | ∞ | ∞ | ∞ |
| 1.0% | ∞ | 3.8% | 0.0% | -5.2% | -1.9% | ∞ | 2.3% | -3.8% | ∞ |
| ∞ | ∞ | 12.3% | 10.8% | 0.0% | ∞ | ∞ | ∞ | ∞ | ∞ |
| ∞ | 3.3% | 4.5% | ∞ | -2.0% | 0.0% | 6.9% | ∞ | ∞ | ∞ |
| ∞ | ∞ | 5.4% | ∞ | -5.4% | -2.0% | 0.0% | ∞ | ∞ | ∞ |
| 2.1% | ∞ | 5.3% | 1.5% | -4.1% | -0.8% | ∞ | 0.0% | -2.6% | ∞ |
| 9.5% | 11.5% | 9.0% | 14.4% | 2.5% | 6.7% | 15.7% | ∞ | 0.0% | 15.1% |
| 2.9% | -2.0% | 6.6% | 3.8% | -4.3% | 1.5% | 3.2% | 2.6% | -1.6% | 0.0% |

**Table 3.** Relative $Q$ values of the optimal models on the different 2D samples. Most deviations are within 10%, as it can be expected from a variation of 100 data points on a smooth PDF, but some are infinite. Those are traced back to individual points, far from the training samples', outside the linear approximation range of the bit requirement calculation.

|  | testing sample | | | | | | | | |
|---|---|---|---|---|---|---|---|---|---|
| 0.0% | 0.3% | 1.5% | 3.5% | -4.2% | -1.5% | 4.3% | 7.5% | -5.4% | 5.8% |
| 12.6% | 0.0% | 25.5% | 9.0% | 7.1% | 13.9% | 10.7% | 14.4% | 7.2% | 18.0% |
| 1.7% | 5.0% | 0.0% | 7.8% | -3.9% | -0.7% | 8.4% | 10.2% | -6.0% | 8.2% |
| 1.2% | -4.1% | 4.2% | 0.0% | -5.4% | -1.9% | 0.9% | 1.9% | -4.8% | 2.7% |
| 10.0% | 4.1% | 12.8% | 10.6% | 0.0% | 6.7% | 10.9% | 12.6% | 2.4% | 11.2% |
| 3.4% | 4.4% | 4.6% | 7.9% | -2.1% | 0.0% | 7.9% | 11.3% | -3.7% | 9.4% |
| 1.4% | -5.0% | 5.3% | 0.3% | -5.9% | -2.1% | 0.0% | 1.5% | -4.4% | 1.3% |
| 2.5% | -3.6% | 5.8% | 1.6% | -4.2% | -0.8% | 2.1% | 0.0% | -3.1% | 2.1% |
| 11.2% | 13.6% | 10.5% | 17.1% | 3.5% | 7.9% | 18.1% | 20.7% | 0.0% | 17.2% |
| 3.1% | -2.1% | 7.2% | 4.1% | -4.6% | 1.5% | 3.4% | 2.6% | -2.1% | 0.0% |

**Table 4.** Relative entropy of the optimal models on the different 2D samples. The entropies or log likelihoods of the trained model on the new samples are all finite, unlike some $Q$ values in table 3., showing that the data points in the test samples are not the source of the infinities, but the $\Delta m$ ranges are inaccurate due to the linear approximation.

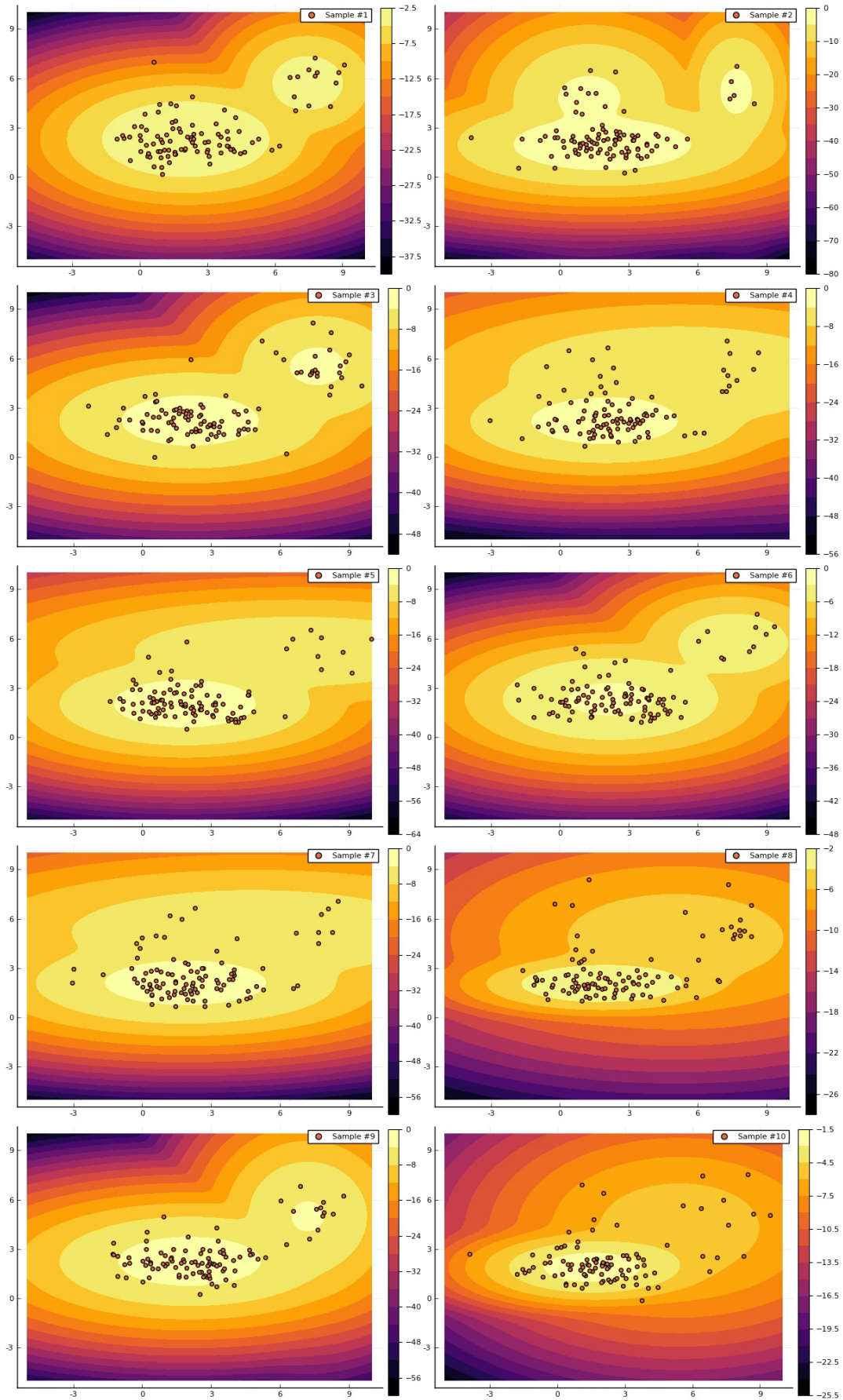

**Figure 3.** The figures show the optimal Gaussian composition for ten different samples from the same 2D distribution. The fitting procedure allowed a maximum of five Gaussians, but the optima contains onlt two or three significant peaks.

Fitting the two dimensional samples results in similarly shaped and smooth outputs, with mostly two significant peaks except for one sample, three peaks, as can be seen on fig. 3. What is surprising is the large amount of infinities in the $Q$ on the testing samples in table 3. As the relative entropy is finite, we can be sure the reason for this is not that the predicted probability is zero at a data point. Instead, the used parameterization $m_k$, and the $\Delta m_k$ ranges don't allow any truncation playground for at least one $u_{\nu i}$ encoded data point. This can either mean that the $\Delta m_k$ regions are too wide, and need to be shrunken according to the new information, or that the new data point falls out of the range assumed for the linear approximation of $F_\nu(x_\mu)$ in the training. Strictly speaking, the model assumes that the encoding of new data points would be done using the linear approximation of $F_\nu(x_\mu)$ nearby a data point in the training sample, and the truncation regions are calculated along those approximations. (Note: It may seem paradoxical that the larger $\Delta x_{\mu i}$ range requires smaller $\Delta m_k$, but only when compared to the used linear appoximation.) The proper treatment of a new data point therefore requires the modification of the $m_k$ parameters and/or the shrinkage of the $\Delta m_k$ truncation regions (making them more precise). In fact, adding a problematic point to a sample only requires a small modification on the $\Delta m_k$ auxiliary parameters to make it valid again, and the new $Q$ value isn't drastically different. Finally, just as was seen for the univariate case, fig. 4 shows that the two dimensional fits stop at three significant peaks when more data is added.

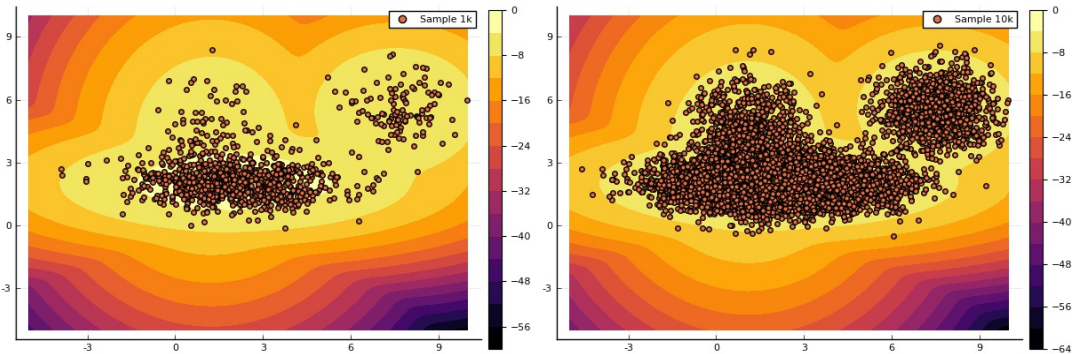

**Figure 4.** Applying the fitting method or more data won't change the characteristic of the resulting model. It gives three significant peaks for 1k and for 10k events.

## 6.3 Known Issues

The code, although capable of fitting samples with arbitrary number of dimensions, it was only tested thoroughly for the uni- and bivariate case. The Gaussian model is also not general, the covariance matrices are restricted to be diagonal.

As the used gradient-free optimization methods require boundaries, some are hard-coded into the wrapper routine. However, the optimization may result in parameters outside the boundaries, due to the soft handling of these limits. When such result is passed to the next optimizer in the chain, it would drop an unhandled exception, a forced termination message.

The code contains functions to prune the model. It was originally intended to remove the Gaussians with small amplitudes, but is also possible to remove Gaussian components with extremely small widths. This is because it was found, that although thin Gausses still contribute to the PDF, the used optimization algorithms seemingly leave those components unaffected. The reason for this is, that the probability density is large around thin Gaussian peaks (e.g. for sigma being $10^{-8}$), and a single peak dominates the PDF value. The probability at the peak is the determinant of the $x_\mu$ derivative of the $F_\nu(x_\mu, m_k)$ modeling function, but that matrix is exactly the negative sum of the derivatives by the Gaussian $a_\mu$ mean parameters. So, for single peaks the matrix multiplication $\left(\frac{\partial F_\nu}{\partial x_\mu}\right)^{-1} \frac{\partial F_\nu}{\partial a_\lambda} \frac{\Delta a_\lambda}{\Delta x_\mu}$ term in $Q$ simplifies to a near unit matrix, and allows the $\Delta a_\lambda$ to be large, despite the large $\frac{\partial F_\nu}{\partial a_\lambda}$ term. Although the runaway width problem mainly affects points that could be outliers, it is generally an undesirable model. Fortunately it is easy to detect it and mark such models invalid.

As with any proper PDF, the probabilities need to add up to one. The sample code uses the methods mentioned in Section 3.2. The first leaves the amplitudes of the Gaussians as independent parameters, while performs the normalization internally. Though it is a simple and valid way to address normalization, it makes model comparison ambiguous. A set of probability amplitudes can be expressed in multiple ways, thus having multiple possible $Q$ values for the same PDF. Although a normalization on the amplitude related parameters is possible after the optimization, there are no fundamentally preferred ways to do it. This means, that $Q$ values of this implementation should be attenuated with a non-statistical error, that can be approximated by the amplitude-related $\log \Delta m_k$ values. The less ambiguous way would be addressing the probability normalizations by directly removing the degree of freedom of an amplitude, e.g. by parameterizing the surface where the sum of amplitudes is constant.

For this reason, an a second version of amplitude parameterization was also implemented. It is based on the fact, that the square sum length of vectors don't change under rotations. This may seem rather unusual, as there are no independent amplitudes for each Gaussian component. Nevertheless, this is a more honest representation of the fact that probabilities must add up to one.

As it was noted in the Sections 2.1 and 5, using $Q$ for the comparison of models with different number of degrees of freedom is problematic due to the singularities appearing when $\Delta x \to 0$. The runs using the Julia code showed, that in this case the $\Delta m_k < 1$ constraint prevents such problems, and the $Q$ value for a model with only a few significant peaks is the same as the pruned model's $Q$ value. Nevertheless, it could be a problem in other implementations.

## 7 Nonlinear Extension

The method presented in this article is based on the total derivative of a parametrized encoding map. It was shown, that in the final formulas one does not need to specify the $\Delta x_{\mu i}$ precision goals of the encoded data, it is enough to optimize for the $\frac{\Delta m_k}{\Delta x_{\mu i}}$ ratios or related quantities. However, as it was noted in Sections 4 and 6.2, it would be beneficial to include higher order derivatives of the $F_\nu(x_\mu, m_k)$ map to have control over the uncertainty of the fitted parameters.

The total derivative of eq. 2 can be extended by the Taylor expansion as

$$\mathrm{d}u = \frac{\partial F_\nu}{\partial x_\mu}\Delta x_{\mu i} + \frac{\partial F_\nu}{\partial m_k}\Delta m_k + \frac{1}{2}\frac{\partial^2 F_\nu}{\partial x_\mu \partial x_\lambda}\Delta x_\mu \Delta x_\lambda + \frac{1}{2}\frac{\partial^2 F_\nu}{\partial m_k \partial m_l}\Delta m_k \Delta m_l + \frac{\partial^2 F_\nu}{\partial x_\mu \partial m_k}\Delta x_\mu \Delta m_k + \cdots$$

As long as only the first order $x_\mu$ derivative is kept, the higher order derivatives of $m_k$ might be handled as a single vector $\left(\frac{\partial F_\nu}{\partial x_\mu}\right)^{-1}(F_\nu(x_{\mu i}, m_k + \Delta m_k) - F_\nu(x_{\mu i}, m_k))$, and used as a single determinant perturbing vector in eq. 9. However, the requirement is, that $F_\nu$ for a given $\nu$ index takes its extrema within the $\Delta m_k \in M = \bigcup_k [0, \Delta m_k^{\max}]$ at $\Delta m_k = 0$ and $\Delta m_k^{\max}$. For a linear approximation it is easy to find the directions that decrease the $\left|\det\left(\frac{\partial F_\nu}{\partial x_\mu}\right)\right|$ volume by using the absolute values of each $m_k$ term, as in eq. 9.

The general non-linear approach can be outlined following the observations in Section 2. For each $x_{\mu i}$ data point one needs to define a precision goal, a volume[3] $V_i$ within $x_{\mu i}$ is truncated to. The image of $V_i$ in $u$-space is symbolically $U_i(m_k) = \{F_\nu(x, m_k) | x \in V_i\}$, and has dependence on the $m_k$ parameters. In case the $m_k$ parameters are infinitely precise, the truncation of a $u_{\nu i}$ encoded data point within the $U_i$ volume would give a decoded $x_{\mu i}$ within the precision volume $V_i$. Any truncation of the $m_k$ parameters shifts the available $U_i$ volume, as now the map from $u_\nu$ to $x_\mu$ is not the exact inverse of the $F_\nu(x_\mu, m_k)$ transformation. In case the truncation of $u_{\nu i}$ and $m_k$ happens independently, the points of $U_i(m_k + \Delta m_k)$ shifted out from the $U_i(m_k)$ volume must be removed from the volume of possible $u_i$ truncation target. This means, that any truncation of the $m_k$ parameters shrink the available volume of $u_\nu$. Thus, to keep the reconstruction precision of $x_\mu$ and drop away bits of $m_k$, one has to include more bits of $u_\nu$.

---

[3]. Truncation usually happens in a range, and not in a volume. However, for correlated data sources, like a multidimensional data point, one can sacrifice the precision of a data point chunk to the others, and that behaves like a volume.

With a measure $\varphi(u_\nu, m_k)$, constant one for the $U_i$ volume and zero otherwise, the measure after a $\Delta m_k$ parameter change is $\varphi(F_\nu(x_\mu, m_k + \Delta m_k))$. To make sure a point is available for truncation, one needs to check that for no $\Delta m_k$ it leaves the $U_i$ volume:

$$\prod_{\Delta m_k \in M} \varphi(F_\nu(x_\mu, m_k + \Delta m_k))$$

To remove the dependence of $x_\mu$, one can use the inverse transformation $x_\mu = F_\mu^{-1}(u_\nu, m_k)$. The volume of the region available for $u_\nu$ truncation is the integral of this measure over full $u_\nu$-space (which is usually bounded to $0...1$ in $\mathbb{R}^n$).

This symbolic formula shows, that in the general approach, the calculation of the bit requirement of a $u_\nu$ encoded data point with regard to the $m_k$ truncation requires a back and forth transformation, to see if a data point were pushed out of the allowed region when an independent truncation was done on the $m_k$ parameters. In cases where $F_\nu$ is strictly monotonic for all $m_k$ parameters, and for convex $V_i$ volume, usually one can replace the $\prod_{\Delta m_k \in M} \varphi$ with the $\varphi$ at the boundaries of the $M$ truncation volume.

$M$ might be non-rectangular volume, but the current method does not allow a combined $M$-$U_i$ volume (a non-independent truncation of $u_\nu$ and $m_k$, e.g. truncating $u_\nu$ with the knowledge of the truncated $m_k$). A technical problem is, that a parameter usually affects more than one data point, and encoding the bits of a single data point into a single $m_k$ would just mean the exchange of the function of $m_k$ with $u_{\nu i}$, that $m_k$ is a parameter and not an encoded data point. Furthermore, it depends on the value of the $m_k$ parameters, which $x_{\mu i}$ data points can be grouped together with which $m_k$ parameters. Without that knowledge, it is impossible to say if a precision of a parameter could be sacrificed for the precision of a data point. There is no obvious advantage of mixing the purpose of data fields, a parameter with an encoded data point. Data points might be removed, added later, and the model might be extended with new features. Even with these, it is not impossible to construct methods which combine $M$-$U_i$ volumes, but they certainly make it complicated. All in all, it is possible to construct methods for a combined, non-trivial $M$-$U_i$ truncation volume, but as the typical goal is to simplify the model, a single parameter will depend on several of the data points, the expected gain is a fraction of the $M$ volume, which is typically a small contribution on the logarithmic scale.

One can also argue, that the non-linear terms only matter in the case when $\Delta x$ is defined to be finite (as the infinites in table 3. shows), at least when $F_\nu$ is differentiable, and could be neglected for the $\Delta x \to 0$ case. Finite terms, like the bits of the $m_k$ parameters before the decimal point should be neglected in that scenario. However, as Section 4 shows, there is a natural order of magnitude for $\Delta x_{\mu i}$, that could be used for the uncertainty estimation of the $m_k$ parameters.

## 8 Conclusions

The article explores the possibility of using the bit count of a probabilistic model as a fitting criteria on a statistical sample. It succeeds by constructing a parametric coordinate transformation, whose Jacobi determinant is related to the probability density. However, the formula for bit requirement of the model requires not only the spatial derivatives, but the total derivative of the transformation and an approximation using the matrix determinant lemma. As it turns out, the total bit requirement is a perturbation on the negative log likelihood, and minimizing it is equivalent to a regularized maximum likelihood fit. In fact, the regularization favors fewer number of parameters, especially fewer amplitude-components of the probability density estimator, and smaller derivatives, which leads to smoother modeling functions. The method was implemented for a Gaussian mixture model, giving very similar, smooth and low degree of freedom outputs for the different samples from the same distribution. The method is applicable to most regression tasks, as those can be turned into a marginal probability distribution modeling. It is clear that the method can be implemented for some distributions with discontinuities (e.g. uniform), and possibly discrete distributions with real parameters as well (e.g. a mixture of Poissonians), but these require further research.

## 9 Acknowledgements

I've pursued the idea of using the data compression or bit requirement as a fitting criteria, because I believed some element is missing from modern machine learning algorithms. I think I've found it, as the need for data compression, model simplification and comparison are not unique for data science, but for science in general. The development took a significant amount of my time, and I'd like to thank for my wife and daughter for their patience, and for being who they are.